%% file: emnlp2021.tex
\pdfoutput=1

\documentclass[11pt]{article}

\usepackage{emnlp2021}

\usepackage{times}
\usepackage{latexsym}

\usepackage[T1]{fontenc}

\usepackage[utf8]{inputenc}
\usepackage{microtype}
\usepackage{times}
\usepackage{url}
\usepackage{algorithm}
\usepackage{algorithmic}
\usepackage{latexsym}
\usepackage{microtype}
\usepackage{graphicx}
\usepackage{subfigure}
\usepackage{booktabs}
\usepackage{amsmath}
\usepackage{amsthm}
\usepackage{enumitem}
\usepackage{color,amsfonts}
\usepackage{multirow}
\usepackage{tikz}
\usepackage{pgfplots}
\usepackage{cleveref}

\input{notation}

\usepackage{xspace} 
\newcommand{\guda}{\textsc{GUDA}\xspace}
\newcommand{\Dold}{\mathcal{D}_{\textrm{old}}}
\newcommand{\Dnew}{\mathcal{D}_{\textrm{new}}}
\newcommand{\Xnew}{\mathcal{X}_{\textrm{new}}}
\newcommand{\Ynew}{\mathcal{Y}_{\textrm{new}}}
\newcommand{\DT}{\mathcal{D}_{t}}
\newcommand{\DS}{\mathcal{D}_{s}}

\newcommand{\vg}{\pmb{g}}
\newcommand{\vpsi}{\pmb{\psi}}

\title{Generalised Unsupervised Domain Adaptation of Neural Machine Translation with Cross-Lingual Data Selection}

\author{Thuy-Trang Vu \and Xuanli He  \and Dinh Phung \and Gholamreza Haffari \\
     Department of Data Science and AI \\ Faculty of Information Technology,  Monash University, Australia \\
     \texttt{\{trang.vuthithuy,xuanli.he1,first.last\}@monash.edu}
      }

\begin{document}
\maketitle
\begin{abstract}
%
This paper considers the unsupervised domain adaptation problem for neural machine translation (NMT), 
where we assume the access to only monolingual text in either the source or target language in the new domain.
We propose a cross-lingual data selection method to extract  in-domain sentences in the missing language side from a large generic monolingual  corpus.
%
Our proposed method trains an adaptive layer on top of multilingual BERT by contrastive learning to align the representation between the source and target language. This then enables the transferability of the domain classifier between the languages in a zero-shot manner. 
Once the in-domain data is detected by the classifier, the NMT model is then adapted to the new domain by jointly learning translation and domain discrimination tasks.
%
We evaluate our  
cross-lingual data selection 
method on NMT across five diverse domains in three language pairs, as well as a real-world scenario of translation for COVID-19. The results show that our proposed method outperforms other selection baselines up to +1.5 BLEU score.

\end{abstract}

\input{1-intro}

\input{2-problem}
\input{3-method}

\input{4-exp-dataset}

\input{4.1-exp-result}

\input{5-analysis}

\input{6-relatedwork}

\input{7-conclusion}

\section*{Acknowledgments}
This material is based on research sponsored by the ARC Future Fellowship FT190100039; the Air Force Research Laboratory and DARPA under agreement number FA8750-19-2-0501. The U.S. Government is authorized to reproduce and distribute reprints for Governmental purposes notwithstanding any copyright notation thereon. The  authors  are  grateful  to  the  anonymous  reviewers for their helpful comments. The computational resources of this work are supported by the Multi-modal Australian ScienceS Imaging  and  Visualisation  Environment (MASSIVE)\footnote{\url{www.massive.org.au}}.

\bibliography{guda}
\bibliographystyle{acl_natbib}

\appendix
\input{8-appendix}


\end{document}

%% file: notation.tex
\newcommand\vx{\pmb{x}} \newcommand\vy{\pmb{y}}
\newcommand\vz{\pmb{z}}

\newcommand\vh{\pmb{h}}  
  \newcommand\vc{\pmb{c}}

\newcommand\vtheta{\pmb{\theta}} \newcommand\vphi{\pmb{\phi}}

\newcommand\ngram{\mathcal{G}}

%% file: 1-intro.tex
\section{Introduction}


Unsupervised domain adaptation (UDA) aims to generalise  MT models trained on domains with typically large-scale bilingual parallel text to new domains without parallel data \cite{chu-wang-2018-survey}. 
Most prior works in UDA of NMT assume  the availability of either non-parallel texts of both languages or only the \emph{target}-language monolingual text in the new domain to adapt the NMT model. 
The adaptation is achieved by modifying the model architecture and joint training with other auxiliary tasks \cite{gulcehre2015using, domhan-hieber-2017-using, dou-etal-2019-unsupervised}, or
 constructing a parallel corpus for the new domain  from a general-domain parallel text using data-selection methods \cite{silva-etal-2018-extracting,hu-etal-2019-domain}. 
 However, very little attention has been paid to the UDA problem with only the \emph{source}-language monolingual text in the new domain.
 In practice, this setting is not very rare, e.g. building a translation system from English to Shona (a low-resource African language) in a specific domain such as healthcare and disaster. While it would be very time consuming to collect in-domain text in Shona, English corpora are more accessible.




In this paper, we consider the \emph{generalised} problem of UDA in NMT where we  assume the availability of  monotext in only one language, either the source or target, in the new domain. 
We propose a generalised approach to the problem using cross-lingual data selection to extract sentences in the new domain for the missing language side from a large monolingual generic corpus.
Our proposed data selection method trains an adaptive layer on top of multilingual BERT by contrastive learning \cite{pmlr-v119-chen20j}, such that the representations of source and target language are aligned. The aligned representations enable the transferability of a domain classifier trained on one language side to the other language for in-domain data detection. 
Previous works have explored filtering data of the same language for MT \citep{moore-lewis-2010-intelligent, axelrod-etal-2011-domain, 
duh-etal-2013-adaptation, 
junczys-dowmunt-2018-dual}; however, utilising data in one language to detect in-domain data in the other language is under-explored.





With selected sentences in the new domain of the missing language side, the original adaptation problem is transformed to the usual setting of UDA problem, and can be approached by the existing UDA methods. 
In this paper, we extend the discriminative domain mixing method for supervised domain adaptation~\citep{britz-etal-2017-effective} which jointly learns domain discrimination and translation to the unsupervised setting. More specifically, the NMT model jointly learns to translate with the translation loss on pseudo bitext, and captures the characteristics of the new domain by the domain discrimination loss on data from the old and new domains.

Our contributions can be summarised as follows:
\begin{itemize}
    \item We introduce a generalised UDA (\guda) problem for NMT which unifies both the usual setting of having only target language monotext and the under-explored setting with only source language monotext in the new domain.
    \item We propose a cross-lingual data selection method to address \guda by retrieving in-domain sentences of the missing language from a generic monolingual corpus.
    \item We augment the discriminative domain mixing method to UDA by constructing an in-domain pseudo bitext via forward-translation and back-translation. 
    \item We empirically verify the effectiveness of our approach on translation tasks across five diverse domains in three language-pairs, as well as a real-world translation scenario for COVID-19. The experimental results show that our method achieves up to +1.5 BLEU improvement over other data selection baselines. The visualisation of the representations generated by the adaptive layer demonstrates that our method is not only able to align the representation of the source and target language, but it also preserves characteristics of the domains in each space\footnote{Source code is available at \url{https://github.com/trangvu/guda}.}.
\end{itemize}

%% file: 2-problem.tex
\section{Generalised Unsupervised Domain Adaptation}

Domain adaptation is an important problem in NMT as it is very expensive to obtain training data that are both large and relevant to all possible domains. Supervised adaptation problem requires the existence of out-of-domain (OOD) bitext and in-domain bitext. Unsupervised domain adaptation problem assumes OOD and in-domain monotext, usually in the target language.

A domain is defined as a distribution $P(X,Y)$ where $X$ ranges over sentences in the \emph{source} language $s$, and $Y$ is its
translation in the \emph{target} language $t$.
We define the generalised unsupervised domain adaptation (\guda)
for NMT as the problem of adapting an NMT model trained on an old domain $P_{old}(X,Y)$ to a new domain $P_{new}(X,Y)$, where only \emph{either} the source \emph{or} target language text is available in the new domain. 
Since $P(X,Y) = P^s(X)P^{s,t}(Y|X)$, let us consider $P^s_{old}(X)$ which is the distribution over sequences on the source language $s$ in the old domain. It is usually much richer (i.e., containing diverse categories such as news, politics, etc.) than  $P^s_{new}(X)$ which is typically a much more specific category where we aim to adapt the NMT model.
The conditional distribution $P^{s,t}_{old}(Y|X)$ specifies 
the encoder-decoder NMT network to be adapted to the new domain.
%

Given parallel bitext $\Dold=\{(\vx_i,\vy_i)\}$ in the old domain, we consider two settings in \guda:
\begin{itemize}
    \item An initial monolingual text $\Xnew=\{\vx_j\}$ of the source language in the new domain and a  \emph{generic} monolingual text $\DT$ of the target language.
    \item An initial monolingual text $\Ynew =\{\vy_k\}$ of the target language in the new domain and a  \emph{generic} monolingual text $\DS$ of the source language.
\end{itemize}
Crucially, 
in both cases we {\bf do not} require any parallel text in the new domain, hence the term \emph{unsupervised domain adaptation}. 
The goal is to adapt an NMT model, parametrised by $\pmb{\theta}$, trained on the old domain  bitext $\Dold$ to the new domain. 


\begin{figure*}[]
    \centerline{\includegraphics[scale=0.55]{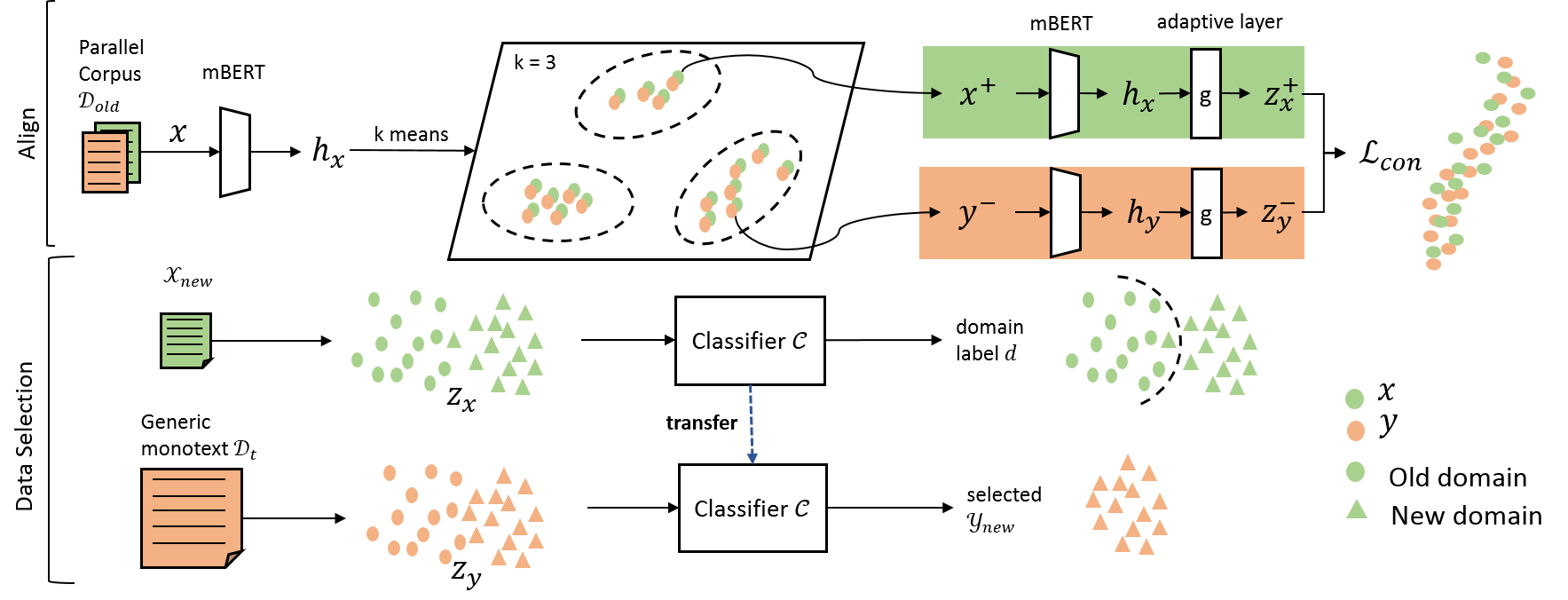}}
    \caption{Our proposed cross-lingual data selection method for \guda with source monotext $\Xnew$.}
    \label{fig:nmt-contrastive}
\end{figure*}

In the  setting involving $\Ynew$, it can be used to create pseudo-parallel data via back-translation \cite{sennrich-etal-2016-improving}, or to adapt the decoder via multi-task learning 
\cite{gulcehre2015using,domhan-hieber-2017-using}. This setting is the usual formulation
in UDA for NMT~\cite{chu-wang-2018-survey}. In contrast, the setting involving the source monotext  $\Xnew$ is not well explored in the literature.

Our approach for addressing \guda is to create \emph{in-domain} monotext for the language side, where the data in the new domain is missing. That is, if given $\Xnew$, we build a \emph{classifier} to select in-domain monotext $\Ynew$ in the target language from the generic monotext $\DT$.
%
We perform a similar procedure for the other case where only in-domain $\Ynew$ is present.
We then adapt the NMT model based on the bitext from the old domain as well as the source and target language monotext in the new domain.
%
The challenge, however, is how to train a classifier for data selection for the language-side with missing data.  
%
We address this problem in Section~\ref{sec:dataselection},  then mention how to adapt the NMT model to the new domain in Section~\ref{sec:uda-nmt}.

%% file: 3-method.tex
\section{Cross-lingual In-domain Data Selection}
\label{sec:dataselection}
\citet{aharoni-goldberg-2020-unsupervised} have shown that the emergent domain clusters via BERT \citep{devlin-etal-2019-bert} can be used to select in-domain \emph{bitext} for NMT. Inspired from that observation, 
we leverage the sentence representations produced by the 
multilingual BERT (mBERT)
for cross-lingual \emph{monotext} selection. We first align the source and target language 
representation space while preserving the domain clustering characteristics in each space.
Using the available monotext in one language, we train a binary classifier to detect old and new domains on the aligned semantic spaces. This classifier is then transferred to pick in-domain sentences in the 
other language (\cref{fig:nmt-contrastive}).

\paragraph{Representation Alignment.} We encode the representation of a sentence $\vx$ by $\vh(\textrm{mBERT}(\vx))$, where $\vh$ computes the mean-pooled top-layer hidden states obtained from mBERT. 
To align the representation space of the source and target language, we learn an \emph{adaptive} layer $\vg_{\vphi}(.)$, a feed-forward network parametrised by $\vphi$, on top of the mBERT by contrastive learning \cite{pmlr-v119-chen20j}. The intuition is that the representation of a translation pair $(\vx_i, \vy_i)$ should be close to each other in the semantic space, {while the representation of non-translation pairs should be far apart.} Specifically, we aim to optimise a contrastive loss,
\begin{align}
\textstyle \mathcal{L}_{\textrm{con}}(\vz^+_{\vx},\vz^+_{\vy}) = -\log \frac{\exp(\textrm{sim}(\vz^+_{\vx},\vz^+_{\vy}))/\tau}{\sum \exp(\textrm{sim}(\vz^+_{\vx},\vz^-_{\vy}))/\tau}
\end{align}
where $\vz_{\vx} := \vg_{\vphi}(\vh(\textrm{mBERT}(\vx)))$ and $\vz_{\vy} := \vg_{\vphi}(\vh(\textrm{mBERT}(\vy)))$ are the output of the adaptive layer for the source and target sentences; $(\vz^+_{\vx},\vz^+_{\vy})$ and $(\vz^+_{\vx},\vz^-_{\vy})$ denote the positive and negative example pairs, $\tau$ is a temperature parameter, and $\textrm{sim}(.)$ is the cosine similarity following \citet{aharoni-goldberg-2020-unsupervised}.
While training $\vphi$ of the adaptive layer, other layers including embedding and transformer layers are frozen. 


Given a batch of N training examples from the old domain $(\vx_i,\vy_i)_{i=1}^{n} \sim \Dold$,
these translation pairs from the bitext are the positive examples. 
Instead of blindly treating 
those
from non-translation pairs as negative examples, we create domain labels by clustering the mBERT representations of the bitext into $k$ clusters. %
For a given $(\vx_i,\vy_i)$ pair
in the training batch, we consider the pairs from distinct clusters in the same batch as the negative examples. This helps the computational complexity by encoding and using all positive and negative examples in the same batch~\citep{pmlr-v119-chen20j}.
We will show the benefit of this setting in \S~\ref{sec:abl_k}. 

\paragraph{In-domain Data Selection.} 
Using the adaptive layer's encoding, we learn a domain classifier for the language-side in which we are given the monotext in the new domain. 
%
Let us assume we are given source language monotext $\Xnew$ in the new domain, and the bitext $\Dold$ in the old domain.\footnote{The other case where we are given $\Ynew$ is similar, and is omitted due to the space constraints.}  
The domain classifier $\vc_{\vpsi}(\vz)$ produces the probability of belonging to the new domain for an input vector 
$\vz$. 
We train the parameter $\vpsi$ for the domain classifier by minimising the following loss (\cref{fig:nmt-contrastive}),
\begin{align}
\begin{split}
\textstyle
\mathcal{L}_{\textrm{disc}}(\vpsi) =&-\sum_{\vx \in \Xnew} \log(\vc_{\vpsi}(\vg(\vh(\vx)))) \\
&- \sum_{\vx \in \Dold^s} \log(1 - \vc_{\vpsi}(\vg(\vh(\vx))))
\end{split}
\end{align}
where $\Dold^s$ denotes source language side of the parallel bitext $\Dold$. 
Thanks to the aligned semantic spaces, we then \emph{transfer} the trained domain classifier cross-lingually to the other language-side to select a subcorpus of in-domain monotext. 
%
We select the top-$k$ probable sentences from the given generic corpus of the other language-side. 

\section{NMT Adaptation to the New Domain}
\label{sec:uda-nmt}

\begin{figure}[]
    \centerline{\includegraphics[scale=0.7]{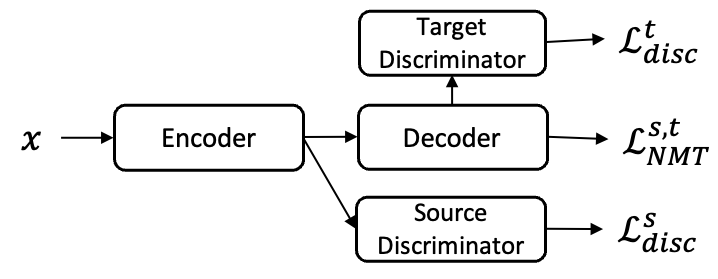}}
    \caption{Discriminative domain mixing approach to UDA for NMT}
    \label{fig:nmt-uda}
\end{figure}

 Given the parallel data in the old domain $\Dold$ and   monolingual data in the new domain  for both the source language $\Xnew$ and target language $\Ynew$, we adapt the NMT model
 by minimising the loss, 
\begin{align}
\mathcal{L} = \mathcal{L}^{s,t}_{\textrm{NMT}} + \mathcal{L}^{{s}}_{\textrm{disc}} +  \mathcal{L}^{{t}}_{\textrm{disc}}   
\label{eq:disc-mixing}
\end{align}
as illustrated in~\cref{fig:nmt-uda} and explained below.

\paragraph{Bitext Loss.} 
We create \emph{pseudo}-bitext $\Dnew$ by back-translating $\Ynew$ using a reverse-direction translation model trained on $\Dold$.
The quality of the pseudo-bitext depends on 
the quality of 
the reverse-direction NMT model in the new domain. 
We further mix the pseudo-bitext $\Dnew$
with the old-domain bitext $\Dold$
to form the bitext loss function 
\begin{align}
\begin{split}
\textstyle
\mathcal{L}_{\textrm{NMT}}^{s,t}(\vtheta)=&-\sum_{(\vx,\vy)\in \Dnew} \log p_{\vtheta}(\vy|\vx) \\
&- \lambda_1   \sum_{(\vx,\vy)\in \Dold} \log p_{\vtheta}(\vy|\vx)
\end{split}
\end{align} 
where $p_{\vtheta}(\vy|\vx)$ is the translation probability according to the NMT model, and $\lambda_1$ controls  effect of the old domain. 

\paragraph{Source Monotext Loss.} To take into account the clean text in source language of the new domain, we apply the discriminative domain mixing method \citep{britz-etal-2017-effective} to force the encoder towards capturing new domain's characteristics. For this purpose, we  build a classifier $\vc_{\vpsi_{e}}(\vz_{e})$, a feedforward network parametrised by $\vpsi_{e}$, whose output is the new domain's probability.  $\vz_{e} = \vh(\textrm{enc}_{\vtheta}(\vx))$ is the representation of the sentence $\vx$, computed by the mean-pooled average of the top layer's states of the NMT's encoder. The source monotext loss
is then defined as,
\begin{align}
\begin{split}
\textstyle
    \mathcal{L}_{\textrm{disc}}^{s}&(\vtheta,\vpsi_e) = - \sum_{\vx\in \Xnew} \log \vc_{\vpsi_e}(\vh(\textrm{enc}_{\vtheta}(\vx))) \\
    &- \lambda_2  \sum_{\vx\in \Dold^s} \log (1-\vc_{\vpsi_e}(\vh(\textrm{enc}_{\vtheta}(\vx))))
\end{split}
\end{align}
where $\lambda_2$ controls the effect of the old domain.

\paragraph{Target Monotext Loss.} Similarly, the target monotext loss 
is defined as,
\begin{align}
\begin{split}
\textstyle
    \mathcal{L}^{{t}}_{\textrm{disc}}&(\vtheta, \vpsi_d) =-\sum_{\vy\in \Ynew} \log \vc_{\vpsi_d}(\vh(\textrm{dec}_{\vtheta}(\vy))) \\
    &- \lambda_3\sum_{\vy\in \Dold^t} \log (1-\vc_{\vpsi_d}(\vh(\textrm{dec}_{\vtheta}(\vy)))
\end{split}
\end{align}
where $\textrm{dec}_{\vtheta}$ is the NMT's decoder, $\vc_{\vpsi_d}$ is the domain classifier parametrised by $\vpsi_d$ for the decoder, $\Dold^t$ is the target sentences in the old domain's bitext, and $\lambda_3$ controls the effect of the old domain.

%% file: 4-exp-dataset.tex
\section{Experiments}
We evaluate our 
proposed approach for \guda on the three language pairs 
covering five domains, and a real-world translation task, namely, TICO-19. 

\subsection{Setup}

\paragraph{Datasets.} Table~\ref{tab:dataset} shows data statistics. The general domain datasets come from  WMT2014 for English-French, WMT2020 for English-German, news parallel corpus from OPUS for Arabic-English\footnote{GlobalVoices, News-Commentary, UN,  WikiMatrix, UNPC}. We appraise our proposed methods on following specific domains: TED talk, Law, Medical, IT, Koran from OPUS~\citep{tiedemann-2012-parallel} following the recipe 
in
\citet{koehn-knowles-2017-six}. 
%
%
We sample 10M English sentences from Newcrawl 2007-2019 as the generic monolingual corpus. 
Data pre-processing is described in~\Cref{apendix:hyperparam}.

\input{tab-dataset}

\input{tab-main-result}

\paragraph{Baselines.} We evaluate the effectiveness of our GUDA framework over the zero-shot baseline (\textbf{base}) where the old-domain model is evaluated without any further training on the new domain. We also evaluate our method against a pseudo-translation baseline (\textbf{trans}) where the old-domain model is further trained on the pseudo-translation of monolingual data from the new domain. More specifically, the pseudo-translation training data contains sentences in the source language and its forward-translated sentences in English for to-English translation direction. Otherwise, it contains sentences in the target language and their back-translated sentences in English for from-English translation direction. We also train fully-supervised models (\textbf{sup.}) which further trains the old-domain models on in-domain parallel data and yields approximately the upperbound BLEU scores. 

We compare our proposed 
in-domain data selection method against several baselines including,
\begin{itemize}
    \item \textbf{random}: we randomly select  English sentences from the generic monolingual pool and treat them as in-domain sentences.
    \item \textbf{cross entropy difference (CED)} \citep{moore-lewis-2010-intelligent} which is a widely used data selection method in MT. 
    The CED score of a given sentence $x$ in the generic corpus  is calculated as $\textrm{CED}(x) = H_S(x)-H_G(x)$, where $H_S(x)$ and $H_G(x)$ are the cross-entropy of the sentence $x$ according to the specific domain and generic domain LMs respectively. The lower the CED score is, the more likely the sentence belongs to this specific domain. In our GUDA setting, to enable cross-lingual data selection, we train a multilingual neural LM on the bitext in the old domain then further finetune it on the available monotext in the new domain and use it to rank the generic corpus. We only run CED methods for En$\leftrightarrow$Fr and En$\leftrightarrow$De translation since we do not share vocabulary between  Ar and En.
    \item \textbf{domain-finetune} \citep{aharoni-goldberg-2020-unsupervised} which trains a domain classifier on mBERT representations and selects the top-k in-domain sentences scored by the classifier. Despite of having similar selection mechanism to our method, the classifier in the domain-finetune technique operates on the 
    pretrained representation space of mBERT without
    alignment between languages.
\end{itemize}

\paragraph{GUDA setup.} We assume the availability of
non-English language data
and evaluate our method to select 500K English sentences from the generic monolingual pool. 
We use the multilingual DistillBERT(mDistillBERT) \citep{sanh2019distilbert} to encode the sentence representation.
We sample and cluster 2M sentences from the old-domain bitext 
into k=5 clusters for negative example creation.
To train the domain classifier, we extract the top 500K sentences from the old domain with low similarity scores between their representation and the mean representation of the monotext in the new domain.


The adaptive layer is a 2-layer feed-forward network with hidden size 128. We set the temperature parameter $\tau$ in the contrastive loss to 0.2. We train the adaptive layer using the Adam optimiser with learning rate 1e-5
, batch size of 64 sentences, 
up to 20 epochs with early stopping if there is no improvement for 5 epochs on the loss of the dev set in the old domain. The domain discriminator is also a 2-layer feed-forward network with the same hyperparameters as the adaptive layer.
We use the Transformer~\citep{transformer} as NMT model and set the mixing hyperparameters $\lambda_1, \lambda_2, \lambda_3$ to 1, i.e. the old domain parallel data as well as source and target monotext contributes equally to the training signal for the NMT model. Detail of the model hyperparameters can be found in the Appendix~\ref{apendix:hyperparam}.


%% file: tab-dataset.tex
\begin{table}[t]
\begin{center}
\scalebox{1}{
\begin{tabular}{l||ccc}
\toprule
\textbf{Domain} &\multicolumn{1}{c}{\textbf{Fr-En}} & \multicolumn{1}{c}{\textbf{De-En}} 
&\multicolumn{1}{c}{\textbf{Ar-En}} \\
\midrule
\textsc{News}  & 35.7M & 40.5M & 21.2M \\
\textsc{Law}   & 625K  & 454K  & -\\
\textsc{Med}   & 689K & 231K  & - \\
\textsc{IT}    & 362K & 158K  & 246K \\
\textsc{Koran} & 128K & 17.8K  &  183K\\
\textsc{TED}   & 190K & 164K  & 199K\\
\bottomrule                 
\end{tabular}
}
\caption{Number of training sentences in the evaluation datasets. Each dataset contains 2K dev and test sentences.}
\label{tab:dataset}
\end{center}
\end{table}


%% file: tab-main-result.tex
\begin{table*}[]
\begin{center}
\scalebox{0.8}{
\begin{tabular}{l||ccccc|ccccc
|ccc}
\toprule
& \multicolumn{5}{c|}{\textbf{Fr-En}} & \multicolumn{5}{c}{\textbf{De-En}}  & 
\multicolumn{3}{|c}{\textbf{Ar-En}}\\
& 
\multicolumn{1}{c}{law} & \multicolumn{1}{c}{med} & \multicolumn{1}{c}{IT} & \multicolumn{1}{c}{Koran} & \multicolumn{1}{c}{TED}  &
\multicolumn{1}{|c}{law} & \multicolumn{1}{c}{med} & \multicolumn{1}{c}{IT} & \multicolumn{1}{c}{Koran} & \multicolumn{1}{c}{TED}  &
\multicolumn{1}{|c}{IT} & \multicolumn{1}{c}{Koran} & \multicolumn{1}{c}{TED}
\\
\midrule
\midrule
\multicolumn{13}{l}{\textit{Translate to English}} \\
base & 42.64 & 37.81 & 28.79 & 7.68 & 34.27 &
       39.37 & 37.97 & 35.66 & 14.08 & 36.55 &
       15.32 & 1.91 & 21.84 \\
trans & 43.33 & 40.70 & 31.15 & 7.94 & 33.60 &
        38.35 & 37.50 & 35.48 & 14.08 & 36.40 & 
        3.91 & 0.23 & 14.61 \\
\midrule
rand & 46.33 & 42.08 & 35.48 & 11.15 & 35.48 & 
       44.32 & 39.81 & 37.53 & 18.52 & 36.92 & 
       16.52 & 6.43 & 20.35 \\
CED & 46.60 & 43.33 & 37.66 & 14.56 & 37.41 & 
      48.03 & 45.00 & 42.72 & 19.20 & 38.46 & 
      - & - & -\\
DF & \textbf{47.92} & 44.06 & 38.79 & 16.34 & 37.48 &
     49.87 & 45.37 & 42.52 & \textbf{21.86} & 39.22 & 
     18.98 & 7.74 & 22.39 \\
our & 47.88 & \textbf{44.36}$^\dagger$  & \textbf{38.89}  & \textbf{17.25}$^\dagger$  & \textbf{38.79}$^\dagger$  &
      \textbf{51.01}$^\dagger$  & \textbf{46.61}$^\dagger$  & \textbf{42.73}$^\dagger$  & 21.45$^\dagger$  & \textbf{39.34} &
      \textbf{20.22}$^\dagger$  & \textbf{10.90}$^\dagger$ & \textbf{22.71}$^\dagger$ \\ 
\midrule
sup. & 49.81 & 53.82 & 63.68 & 19.53 & 41.56 &
      61.02 & 53.38 & 43.42 & 20.98 & 40.19 &
      41.61 & 17.44 & 36.71\\
\midrule
\midrule
\multicolumn{13}{l}{\textit{Translate from English}} \\
base & 23.73 & 25.32 & 20.51 & 5.58 & 35.73 &
       34.60 & 34.52 & 29.35 & 11.23 & 31.32 &
       13.66 & 0.30 & 12.77 \\
trans & 33.71 & 28.82 & 35.28 & 12.91 & 35.11 &
        35.13 & 38.80 & 31.50 & 12.35 & 32.58 & 
        12.65 & 0.89 & 12.83 \\
\midrule
rand & 32.43 & 28.53 & 40.02 & 13.77 & 34.97 & 
       33.86 & 36.47 & 30.29 & 12.57 & 32.77& 
       12.37 & 3.01 & 14.46\\
CED & 33.19	& 29.14 & 40.82 & 14.04 & 35.86 & 
      34.81 & 40.62 & 31.02 & 12.52 & 32.83 & 
      - & - & -\\
DF & 34.63 & 29.99 & 41.09 & 14.97 & 36.18 & 
     35.23 & 41.28 & \textbf{31.93} & 13.19 & 33.69 & 
     14.32 & 6.42 & 15.07 \\
our & \textbf{35.67}$^\dagger$  & \textbf{30.59}$^\dagger$  & \textbf{41.48}$^\dagger$  &  \textbf{16.10}$^\dagger$  & \textbf{37.79}$^\dagger$  & 
      \textbf{35.65}$^\dagger$  & \textbf{42.67}$^\dagger$  & 31.81 & \textbf{13.72}$^\dagger$  & \textbf{33.86}$^\dagger$  & 
      \textbf{14.51}$^\dagger$  & \textbf{7.99}$^\dagger$  & \textbf{16.33}$^\dagger$  \\
\midrule
sup. & 40.95 & 41.09 & 53.24 & 22.72 & 40.47 & 
       46.82  & 46.09 & 34.03 & 14.29 & 34.53 & 
       26.64 & 15.74 & 21.85\\
\bottomrule
\end{tabular}
}
\caption{BLEU score 
of \guda under various selection strategies:
random (\textit{rand}), cross-entropy difference (\textit{CED}), domain-finetune (\textit{DF}), and our cross-lingual data selection. \textit{base}  and \textit{sup.} are the scores of zero-shot and fully supervised on in-domain parallel data. \textit{trans} is the NMT model trained on pseudo bitext where monolingual in-domain data is machine translated in the missing side.
Highest scores of \guda are marked in \textbf{bold}. $^\dagger$~indicates that our method is statistically significant difference to the domain-finetune baseline (p-value $\leq$ 0.05).}
\label{tab:result-to-en}
\end{center}
\end{table*}

%% file: 4.1-exp-result.tex
\subsection{Main Results}
\input{tab-covid}
Table~\ref{tab:result-to-en} presents the result of translations to and from English,  according to \guda with source and target language monotext respectively. 
There is a significant gap between the fully supervised (sup.) and zero-shot (base) scores. 
It can be seen that \guda is able to
reduce this gap, especially when the in-domain data are selected intelligently. Overall, our 
selection method consistently outperforms both the domain-finetune and CED strategy.

We further assess our approach on the translation initiative for COVID-19 task (TICO-19) for En-Fr and En-Ar \citep{tico-19}. The task contains a dev set and a test set of 971 and 2100 sentences. As an emerging domain, there is no training set. We collect additional 49K and 17K in-domain French and Arabic monotext\footnote{\url{https://github.com/neulab/covid19-datashare}}.
As shown in Table~\ref{tab:covid19},
surprisingly, \guda on random selection deteriorates the BLEU score. It is possible that pandemic related words have not appeared often before. Consistent with previous results,
our method outperforms other methods up to +1.2 BLEU score. \begin{figure}[]
    \centerline{\includegraphics[scale=0.3]{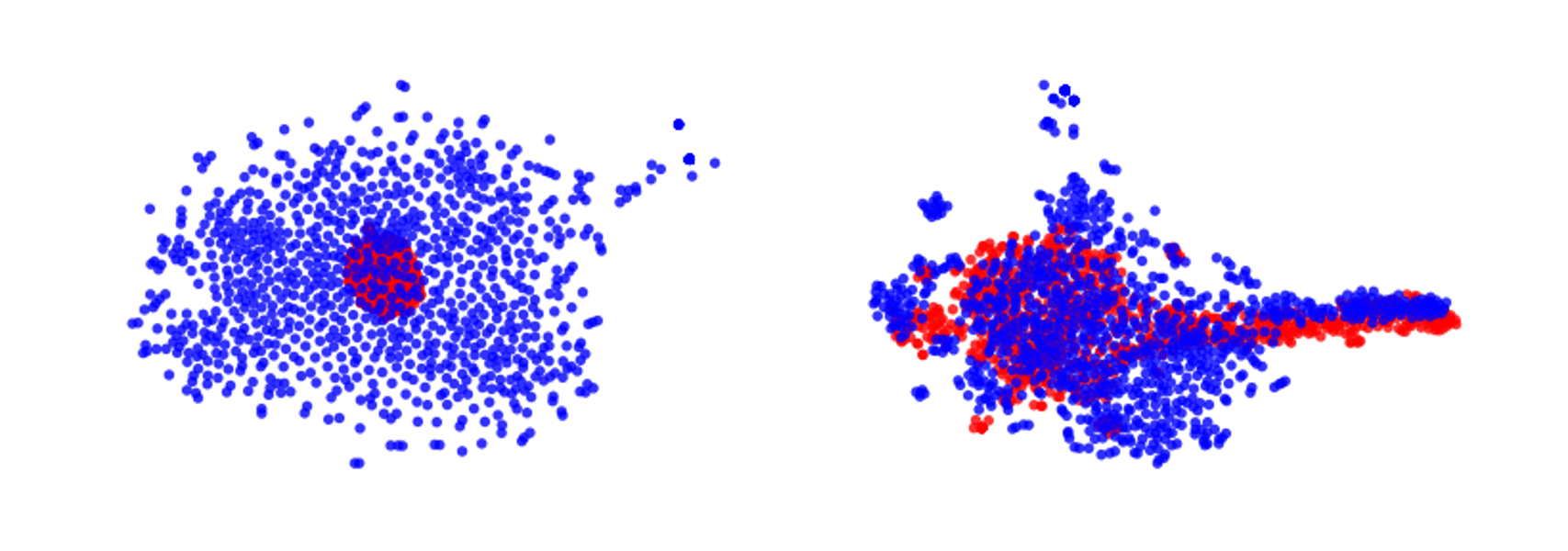}}
    \caption{t-SNE visualisation of the Fr (red) and En (blue) of TICO-19 dev set, encoded by multilingual DistillBERT (left) and the adaptive layer (right).}
    \label{fig:tico-vis}
\end{figure}

To evaluate our alignment method, we visualise
the representation of the TICO-19 dev set produced by mDistillBERT and the adaptive layer in Figure~\ref{fig:tico-vis}. It can be seen that the adapted French and English representations are better aligned in the semantic space than the mDistillBERT.

%% file: tab-covid.tex
\begin{table}[]
\begin{center}
\scalebox{1}{
\begin{tabular}{l||rrrr}
\toprule
 & \textbf{Fr-En} & \textbf{En-Fr} & \textbf{Ar-En} & \textbf{En-Ar}
\\
\midrule
base & 32.35 & 25.07 & 34.11 & 24.52 \\
rand &  30.59 & 24.61 & 32.30 & 24.20\\
CED &  32.55 & 25.13 & - & - \\
DF & 33.25 & 26.24 & 34.56 & 25.36 \\
Our & \textbf{34.17}$^\dagger$ 	& \textbf{27.45}$^\dagger$  & \textbf{35.24}$^\dagger$  & \textbf{26.10}$^\dagger$  \\ 
\bottomrule
\end{tabular}
}
\caption{Results on TICO-19 translation task. $^\dagger$~indicates that our method is statistically significant difference to the domain-finetune baseline (p-value $\leq$ 0.05).}
\label{tab:covid19}
\end{center}
\end{table}

%% file: 5-analysis.tex
\section{Ablation and Analysis}
\subsection{Ablation}
\paragraph{Clustering-based negative sampling.}
\label{sec:abl_k}
The intuition of the clustering-based negative sampling is to preserve the domain clustering characteristics emerged in mBERT. We assess the importance of this clustering step and the effect of the number of cluster $k$ on the En$\leftrightarrow$De translation performance in law, med and TED domains. 
\Cref{tab:k-ablation} reports the BLEU score of the NMT model in the new domain with $k=\{1,2,3,5,7,10\}$ where $k=1$ corresponds to perform negative sampling without pre-clustering mBERT representation space.
\input{tab-abl-k}
\input{tab-disc-mix}
Overall, the NMT model trained on the selected data with clustering-based negative sampling $k>1$ outperforms the one without clustering $k=1$. On the other hand, the effect of number clusters $k$ varies, depending on the domains and languages. From the empirical results, we found that $k=\{5,7\}$ works better than other values.

\begin{figure}[]
    \centerline{\includegraphics[scale=0.3]{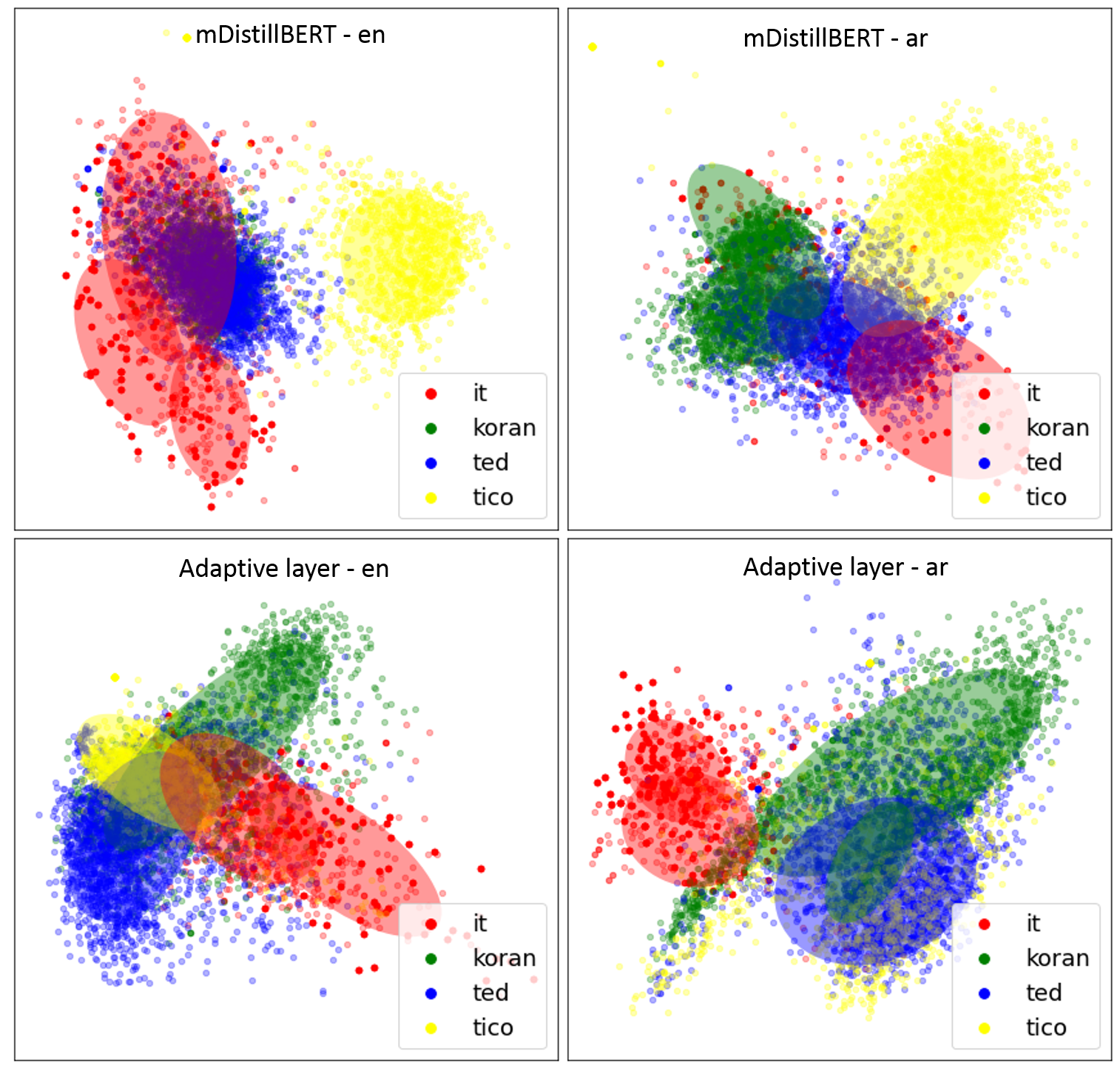}}
    \caption{2D visualisation of the unsupervised GMM-based clustering of En-Ar representations.}
    \label{fig:cluster-visual}
\end{figure}

\paragraph{Discriminative domain mixing.}
We run ablation experiments to verify the contribution of each loss term in the discriminative domain mixing training objective presented in~\cref{eq:disc-mixing}. Particularly, we evaluate the NMT adapted to the new domain using (i) only the bitext loss (\textbf{BI}); (ii) the combination of the bitext loss and either the source monotext loss (\textbf{BI+S}) or the target monotext loss (\textbf{BI+T}); and (iii) the joint of all three loss terms (\textbf{BI+S+T}).
~\Cref{tab:disc-mix} shows the results under both supervised domain adaptation where we have access to the true bitext, and UDA in which the model is trained on pseudo bitext generated by back-translation (warm-start).
The size of the ground-truth bitext is shown in~\Cref{tab:dataset}. The size of the pseudo-bitext is 500K which is approximately double the size of the ground-truth bitext of TED and med domains, and roughly the same for law domain.
We also further evaluate the contribution of the discriminative domain loss when the NMT model is trained from scratch (cold-start).

Consistent with \citet{britz-etal-2017-effective}, training NMT on mixed domain data (BI) degrades performance versus models fit to a single domain (sup.). Adding the discriminative domain loss can mitigate this negative effect in multi-domain NMT.  We observe similar outcomes in both domain adaption with the true bitext and the pseudo bitext. 
 Overall, we found that the source monotext loss plays a more critical role than the target monotext loss. Combining both monotext loss  achieves the best BLEU score in most of domain adaptation scenarios. 



\subsection{Analysis}
\paragraph{Domain cluster visualisation.}
\input{figure-ngram}
\input{figure-cdf}
To demonstrate the ability of our approach in preserving the domain clustered characteristics of mBERT, we plot 2D visualisation of the mean-pooling BERT hidden state sentence representation and our constrastive-based sentence representations using PCA.
Following \citet{aharoni-goldberg-2020-unsupervised}, we combined the development set of all the new domain dataset and cluster the representations using a Gaussian Mixture Model (GMM) with $k$ pre-defined clusters where $k$ is number of domains. 

\Cref{fig:cluster-visual} visualises the obtained clusters in semantic space of mDistillBERT and the adaptive layer for each language in the translation pairs. The ellipses  describe the mean and variance parameters learned for each cluster. In line with the finding in \citet{aharoni-goldberg-2020-unsupervised}, the mDistillBERT representation of English sentences can be clustered by their domains with a small overlap region. In contrast, 
Arabic sentences are not well-clustered according to their domains where their domain clusters exhibit a high overlap rate.
As can be seen, our contrastive-based representation alignment method is not only able to preserve the domain clusters in English sentences but also learn domain clustered representations of Arabic sentences in which the clusters are less overlapped.

\paragraph{Distribution of domain predictive score.}
\Cref{fig:cdf} plots the cumulative distribution for the domain predictive score over the generic English corpus. It can be seen that only a small portion of the generic corpus are predicted to belong to the new domains.  As expected, the more specific-domains such as med and law have smaller number of anticipated sentences than the TED domain.

\paragraph{ngram analysis.} 
A domain can be considerred as a distribution over ngram. The data selection methods mitigate the domain shift in NMT by introducing ngrams of the new domain to the training corpus.
We estimate the new in-domain ngram contribution of each selection method by calculating the overlap of ngrams in the translation hypothesis and the translation reference. 
The new ngram contribution is calculated as 
\begin{align}
\frac{\sum_i\ngram(\Tilde{y}_{i,\textrm{new}}^{\textrm{\guda}}) \cap (\ngram(y_{\textrm{i,new}}^{\textrm{ref}}) \backslash  \ngram(\Tilde{y}_{i,\textrm{new}}^{\textrm{zero}}))}{\sum_i\ngram(y_{\textrm{i,new}}^{\textrm{ref}}) \backslash  \ngram(\Tilde{y}_{i,\textrm{new}}^{\textrm{zero}})}    
\end{align}
where $\ngram(y_{\textrm{i,new}}^{\textrm{ref}}),\ngram(\Tilde{y}_{i,\textrm{new}}^{\textrm{zero}}),\ngram(\Tilde{y}_{i,\textrm{new}}^{\textrm{\guda}})$ are the set of ngrams in the reference, the zero-shot and the \guda translation hypothesis of the sentence $i$ in the test set in the new domain, respectively.

\Cref{fig:ngram} presents the percentage of new ngram contribution, $1\leq n \leq 4$, of each data selection methods as well as the fully supervised model for De-En translation in law, med, ted domains. As expected, the fully-supervised model has the highest correct in-domain ngram rate to the translation hypothesis. Our proposed selection method contributes a higher percentage of in-domain ngrams than other selection methods in all domains. 


%% file: tab-abl-k.tex
\begin{table}[]
\begin{center}
\scalebox{0.9}{
\begin{tabular}{lrrr|rrr}
\toprule
k & \multicolumn{3}{c|}{De-En} & \multicolumn{3}{|c}{En-De}\\
\multicolumn{1}{c}{} & 
\multicolumn{1}{c}{law} & \multicolumn{1}{c}{med} & \multicolumn{1}{c|}{TED}
& \multicolumn{1}{c}{law} & \multicolumn{1}{c}{med} & \multicolumn{1}{c}{TED}
\\
\midrule
1 & 50.76 & 46.32 & 38.61 & 34.98 & 41.45 & 31.92 \\
2 & 50.60 & 46.44 & 38.43 & 35.14 & 41.71 & 33.01  \\
3 & 50.86 & 46.62 & 39.20 &\textbf{ 35.77} & 42.34 & 33.51 \\
5 & 51.01 & 46.61 & 39.34 & 35.65 & \textbf{42.67} & \textbf{33.86} \\
7 &  \textbf{51.04} & 46.62 & \textbf{39.67} & 35.02 & 42.05 & 33.06 \\ 
10 & 50.81 & \textbf{46.70} &  39.39 & 35.21  & 42.24 & 35.35 \\
\bottomrule
\end{tabular}
}
\caption{Cluster-based negative sampling ablation. $k$ is the number of clusters.}
\label{tab:k-ablation}
\end{center}
\end{table}

%% file: tab-disc-mix.tex
\begin{table}[t]
\begin{center}
\scalebox{0.8}{
\begin{tabular}{lrrr|rrr}
\toprule
 & \multicolumn{3}{c|}{De-En} & \multicolumn{3}{c}{En-De}\\
\multicolumn{1}{c}{} & 
\multicolumn{1}{c}{law} & \multicolumn{1}{c}{med} & \multicolumn{1}{c|}{TED} &
\multicolumn{1}{c}{law} & \multicolumn{1}{c}{med} & \multicolumn{1}{c}{TED}
\\
\midrule
\midrule
sup. & 61.02 & 53.38 & 40.19 & 46.82 & 46.09 & 34.53 \\ 
\midrule
\midrule
\multicolumn{7}{l}{\textit{True Bitext}} \\
BI & 53.59 & 49.35 & 40.01 & 
45.69 & 43.10 & 30.78 \\
BI+S & 54.69 & \textbf{51.47} & 40.33  & 
46.98 & 45.59 & \textbf{31.83}  \\
BI+T & 54.70 & 51.31 & 40.28 &
46.84 & 45.62 & 31.63\\
BI+S+T & \textbf{54.73} & 51.38 & \textbf{40.38} &  
\textbf{47.11} & \textbf{45.79} & 31.67 \\
\midrule
\midrule
\multicolumn{7}{l}{\textit{Pseudo Bitext - Warm Start}} \\
BI & 48.57 & 45.62 & 38.61 & 
35.04 & 39.98 & 31.99 \\
BI+S & 50.65 & 46.50 & 39.05 &
35.27 & 41.97 & 33.33 \\
BI+T & 50.22 & 46.27 & 38.88 &
35.16 & 40.68 & 33.16 \\
BI+S+T & \textbf{51.01} & \textbf{46.61} & \textbf{39.34} & 
\textbf{35.65} & \textbf{42.67} & \textbf{33.86} \\
\midrule
\midrule
\multicolumn{7}{l}{\textit{Pseudo Bitext - Cold Start}} \\
BI & 29.33 & 35.02 & 30.83 & 30.98 & 35.28 & 28.19 \\
BI+S & 35.39 & 37.28 & 33.02 & 32.03 & \textbf{37.81} & 30.35 \\
BI+T & 35.68 & 37.13 & \textbf{33.49} & 32.37 & 37.50 & \textbf{30.80} \\
BI+S+T & \textbf{36.07} & \textbf{37.78} & 33.47 & \textbf{33.24}& 37.73 & 30.47  \\
\bottomrule
\end{tabular}
}
\caption{Domain discriminative mixing ablation}
\label{tab:disc-mix}
\end{center}
\end{table}

%% file: figure-ngram.tex
    
    
 
 \pgfplotstableread[row sep=\\,col sep=&]{
    method & rand & CED & DF & our & label \\
1-gram & 10.08 & 22.48 & 28.72 & 35.26 & 43.77 \\
2-gram & 6.57 & 11.93 & 17.88 & 29.45 & 37.05 \\
3-gram & 4.68 & 9.74 & 12.17 & 25.25 & 32.32 \\
4-gram & 03.38 & 15.23 & 17.42 & 22.20 & 28.86 \\
    }\lawngram
    
\pgfplotstableread[row sep=\\,col sep=&]{
    method & rand & CED & DF & our & label \\
1-gram & 8.36 & 20.74 & 20.80 & 26.08 & 34.49 \\
2-gram & 5.86 & 17.46 & 17.49 & 24.58 & 32.88 \\
3-gram & 4.13 & 19.14 & 19.18 & 22.41 & 30.60 \\
4-gram & 3.03 & 17.69 & 17.75 & 20.24 &  28.32 \\
    }\medngram
    
\pgfplotstableread[row sep=\\,col sep=&]{
   method & rand & CED & DF & our & label \\
1-gram & 8.55 & 17.28 & 20.37 & 20.10 & 24.77\\
2-gram &  6.04 & 13.91 & 14.53 & 15.30 & 18.57\\
3-gram &  4.12 & 10.40 & 11.42 & 12.44 & 15.01\\ 
4-gram &  2.80 & 9.43 & 10.22 & 10.27 & 12.30 \\
    }\tedngram
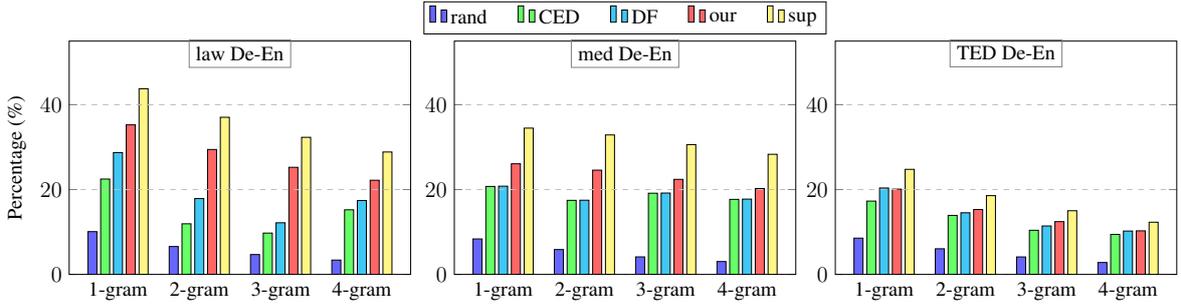
\begin{figure*}[t]
\vspace{-2mm}
\centering 
  \begin{tikzpicture}[thick,scale=0.7, every node/.style={scale=0.7}]
  \tikzstyle{every node}=[font={\fontsize{12pt}{5}}]

      \begin{axis}[
            ybar, axis on top,
            title style={at={(0.5,0.96)},anchor=north,yshift=-0.1, draw=gray},
            title={law De-En},
            height=6cm, 
            width=8cm,
            bar width=0.17cm,
            ymajorgrids, tick align=inside,
            major grid style=dashed,
            enlarge y limits={value=.1,upper},
            ymin=0, ymax=50,
            x tick style={draw=none},
            enlarge x limits=0.2,
            legend style={
                at={(0.5,-0.2)},
                anchor=north,
                legend columns=-1,
                /tikz/every even column/.append style={column sep=0.5cm}
            },
            ylabel={Percentage (\%)},
            ylabel near ticks,
            symbolic x coords={
               1-gram, 2-gram, 3-gram, 4-gram},
           xtick=data,
        ]
        \addplot[fill=blue!60] table[x=method,y=rand]{\lawngram};
        \addplot[fill=green!60] table[x=method,y=CED]{\lawngram};
        \addplot[fill=cyan!60] table[x=method,y=DF]{\lawngram};
        \addplot[fill=red!60] table[x=method,y=our]{\lawngram};
        \addplot[fill=yellow!60] table[x=method,y=label]{\lawngram};
      \end{axis}
      \end{tikzpicture}
  \begin{tikzpicture}[thick,scale=0.7, every node/.style={scale=0.7}]
  \tikzstyle{every node}=[font={\fontsize{12pt}{5}}]
      \begin{axis}[
            ybar, axis on top,
            title style={at={(0.5,0.96)},anchor=north,yshift=-0.1, draw=gray},
            title={med De-En},
            height=6cm, 
            width=8cm,
            bar width=0.17cm,
            ymajorgrids, tick align=inside,
            major grid style=dashed,
            enlarge y limits={value=.1,upper},
            ymin=0, ymax=50,
            x tick style={draw=none},
            enlarge x limits=0.2,
            legend style={
                at={(0.5, 1.17)},
                anchor=north,
                legend columns=-1,
                /tikz/every even column/.append style={column sep=0.5cm}
            },
            symbolic x coords={
               1-gram, 2-gram, 3-gram, 4-gram},
           xtick=data,
        ]
        \addplot[fill=blue!60] table[x=method,y=rand]{\medngram};
        \addplot[fill=green!60] table[x=method,y=CED]{\medngram};
        \addplot[fill=cyan!60] table[x=method,y=DF]{\medngram};
        \addplot[fill=red!60] table[x=method,y=our]{\medngram};
        \addplot[fill=yellow!60] table[x=method,y=label]{\medngram};
        \legend{rand, CED, DF, our, sup}
      \end{axis}
      \end{tikzpicture}
      \hspace{-5.5mm}
  \begin{tikzpicture}[thick,scale=0.7, every node/.style={scale=0.7}]
  \tikzstyle{every node}=[font={\fontsize{12pt}{5}}]
      \begin{axis}[
            ybar, axis on top,
            title style={at={(0.5,0.96)},anchor=north,yshift=-0.1, draw=gray},
            title={TED De-En},
            height=6cm, 
            width=8cm,
            bar width=0.17cm,
            ymajorgrids, tick align=inside,
            major grid style=dashed,
            enlarge y limits={value=.1,upper},
            ymin=0, ymax=50,
            x tick style={draw=none},
            enlarge x limits=0.2,
            legend style={
                at={(0.5,-0.2)},
                anchor=north,
                legend columns=-1,
                /tikz/every even column/.append style={column sep=0.5cm}
            },
            symbolic x coords={
               1-gram, 2-gram, 3-gram, 4-gram},
           xtick=data,
        ]
        \addplot[fill=blue!60] table[x=method,y=rand]{\tedngram};
        \addplot[fill=green!60] table[x=method,y=CED]{\tedngram};
        \addplot[fill=cyan!60] table[x=method,y=DF]{\tedngram};
        \addplot[fill=red!60] table[x=method,y=our]{\tedngram};
        \addplot[fill=yellow!60] table[x=method,y=label]{\tedngram};
      \end{axis}
      \end{tikzpicture}
\caption{Percentage of newly introduced correct ngram.}
\label{fig:ngram}
\end{figure*}

%% file: figure-cdf.tex
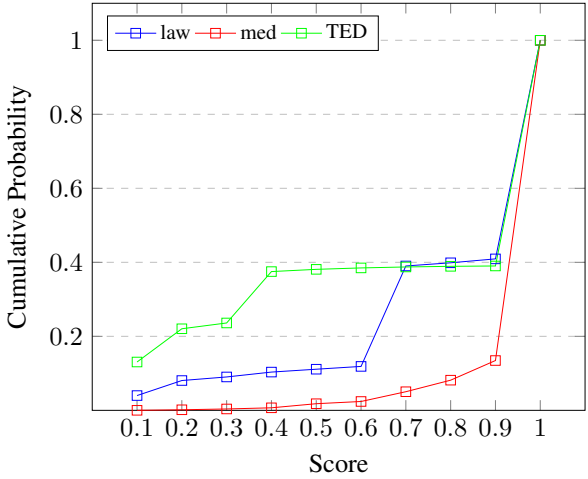
\begin{figure}[]
\centering 
\begin{tikzpicture}[thick,scale=0.9, every node/.style={scale=90}]
  \tikzstyle{every node}=[font={\fontsize{11pt}{4}}]
\begin{axis}[
    xlabel={Score},
    ylabel={Cumulative Probability},
    xmin=0, xmax=1.1,
    ymin=0, ymax=1.1,
    xtick={0.1, 0.2, 0.3, 0.4, 0.5, 0.6, 0.7, 0.8, 0.9, 1},
    ytick={0.2, 0.4,  0.6,  0.8,  1},
    legend pos=north west,
    legend style={nodes={scale=0.8},legend columns=-1,},
    ymajorgrids=true,
    grid style=dashed,
    style={font=\small},
    ylabel near ticks,
    xlabel near ticks,
    width = 250pt,
]
\addplot[
    color=blue,
    mark=square,
    ]
    coordinates {
    (0.1,0.0399587)(0.2,0.0805749)(0.3,0.0902259)(0.4,0.1033982)(0.5,0.1111391)(0.6,0.1186967)(0.7,0.3901544)(0.8,0.398708)(0.9,0.409291)(1,1)
    };
\addplot[
    color=red,
    mark=square,
    ]
    coordinates {
    (0.1, 6.72E-05)(0.2, 0.0015371)(0.3, 0.0037193)(0.4, 0.0068605)(0.5, 0.0179267)(0.6, 0.0239285)(0.7, 0.0503419)(0.8, 0.0815167)(0.9, 0.1346116)(1, 1)
    };
\addplot[
    color=green,
    mark=square,
    ]
    coordinates {
    (0.1, 0.1306227)(0.2, 0.2204599)(0.3, 0.2361131)(0.4, 0.3749861)(0.5, 0.3810024)(0.6, 0.3848688)(0.7, 0.3874596)(0.8, 0.3891659)(0.9, 0.3901408)(1, 1)
    };
\addlegendentry{law}
\addlegendentry{med}
\addlegendentry{TED}
\end{axis}
\end{tikzpicture}
\caption{CDF of predicted score produced by the domain classifier. The smaller the score is, the higher probability the sentence belongs to the new domain.
}
\label{fig:cdf}
\end{figure}

%% file: 6-relatedwork.tex
\section{Related Works}
\paragraph{Unsupervised Domain Adaptation.}
Previous works in UDA has been focused on aligning domain distribution by minimising the discrepancy between representations of source and target domains~\citep{Shen2018, wang-etal-2018-label-aware}; learning domain-invariant representation via adversarial learning~\citep{Ganin2015, shah-etal-2018-adversarial, moghimifar-etal-2020-domain}; bridging the domain gap by adaptive pretraining of contextualised word embeddings~\citep{han-eisenstein-2019-unsupervised,vu-etal-2020-effective}. In this paper, we adapt the NMT model from the old to new domain by learning domain-invariant representations of both encoder and decoder via domain discrimination loss.

\paragraph{Unsupervised Domain Adaptation of NMT.}
There are two main approaches in UDA for NMT, including model-centric and data-centric methods~\citep{chu-wang-2018-survey}. In the model-centric approach, the model architecture is modified and jointly trained on MT tasks, and other auxiliary tasks such as language modelling \cite{gulcehre2015using}. On the other hand, the data-centric methods focus on constructing in-domain parallel corpus by data-selection from general corpus \cite{domhan-hieber-2017-using}, and back-translation \cite{jin2020simple, Mahdieh2020}. Most prior works in UDA of NMT often assume the availability of in-domain data in the target language.
While there are few studies on the UDA problem with in-domain source-language data in statistical MT~\citep{mansour-ney-2014-unsupervised,  cuong-etal-2016-adapting}, this problem remains unexplored in NMT.

\paragraph{Data selection for NMT.} To address the scarcity problem of MT parallel data in specific-domain, data selection methods utilise an initial in-domain training data to select relevant additional sentences from a generic parallel corpus. 
Previous research has used n-gram language model \citep{moore-lewis-2010-intelligent, axelrod-etal-2011-domain,duh-etal-2013-adaptation}, count-based methods \citep{way2018data,parcheta2018data}, similarity score of sentence embeddings \citep{Wang2017SentenceEmbeddingNeural,junczys-dowmunt-2018-dual, Dou2020} to rank the generic corpus. 
The ranking and selection process often operate in the same language, either source or target language, and take advantage of the parallel corpus to retrieve the paired translation~\citep{farajian-etal-2017-multi}. When such generic parallel corpus is unavailable, cross-lingual data selection which uses data in one language to detect in-domain data in the other language is under-explored.


%% file: 7-conclusion.tex
\section{Conclusion}
We have proposed a cross-lingual data selection method to the GUDA problem for NMT where only monolingual data from one language side is available in the new domain. We first learn an adaptive layer to align the BERT representation of the source and target languages. We then utilise a domain classifier trained on one language to select in-domain data for another. Experiments on translation tasks of several language pairs and domains show the effectiveness of our method over other baselines.

%% file: 8-appendix.tex
\clearpage 
\section{Training Procedure}
\label{apendix:hyperparam}
\paragraph{Data preprocessing.} We tokenise English, French, German sentences using Moses tokenizer~\citep{koehn-etal-2007-moses} and remove the sentences with more than 175 tokens. Arabic text are tokenised using {CAM}e{L}~\citep{obeid-etal-2020-camel}. For Arabic, we first filter out the sentences containing more than 50\% Latin characters, then remove those with more than 175 tokens.

\paragraph{Model hyperparameters.} 
The adaptive layer is a 2-layer feed-forward net with hidden size 128. We set the temperature parameter $\tau$ in the contrastive loss to 0.2. We train the adaptive layer using the Adam optimiser with learning rate 1e-5, batch size of 64 sentences, up to 20 epochs with early stopping if there is no improvement for 5 epochs on the loss of the dev set in the old domain. The domain discriminator is also a 2-layer feed-forward net. We train it with the same hyperparameters as in the adaptive layer.

We use the Transformer as NMT model, which consists of 6 encoder and decoder layers, 4 self-attention heads, hidden size of 256,  feed-forward hidden size of 1024, implemented in Fairseq framework \cite{ott-etal-2019-fairseq}. Number of parameters is 64.3M.  We use the Adam optimiser with learning rate 5e-4 \citep{kingma:adam} and an inverse square root schedule with warm-up 1000 steps. We apply dropout and label smoothing with a rate of 0.3 and 0.1 respectively. We learn the vocabulary of size 32000 using unigram language model \citep{kudo-2018-subword}, implemented in SentencePiece\footnote{\url{https://github.com/google/sentencepiece}}. For En-Fr, En-De, and En-Cs, the source and target embeddings are shared and tied with the last layer. We set the mixing hyperparameters $\lambda_1, \lambda_2, \lambda_3$ to 1, i.e. the old domain parallel data as well as source and target monotext contributes equally to the training signal for the NMT model. We train the NMT with the batch size of 32768 tokens and up to 30 epochs with early stopping if there is no improvement on dev set for 5 epochs.

Our model is trained on a V100 GPU, and took up to 4 days for the NMT trained in old domain, and 1 day for other experiments.



%% file: emnlp2021.bbl
\begin{thebibliography}{41}
\expandafter\ifx\csname natexlab\endcsname\relax\def\natexlab#1{#1}\fi

\bibitem[{Aharoni and Goldberg(2020)}]{aharoni-goldberg-2020-unsupervised}
Roee Aharoni and Yoav Goldberg. 2020.
\newblock Unsupervised domain clusters in pretrained language models.
\newblock In \emph{Proceedings of ACL}.

\bibitem[{Anastasopoulos et~al.(2020)Anastasopoulos, Cattelan, Dou, Federico,
  Federman, Genzel, Guzm\'{a}n, Hu, Hughes, Koehn, Lazar, Lewis, Neubig, Niu,
  \"{O}ktem, Paquin, Tang, and Tur}]{tico-19}
Antonios Anastasopoulos, Alessandro Cattelan, Zi-Yi Dou, Marcello Federico,
  Christian Federman, Dmitriy Genzel, Francisco Guzm\'{a}n, Junjie Hu, Macduff
  Hughes, Philipp Koehn, Rosie Lazar, Will Lewis, Graham Neubig, Mengmeng Niu,
  Alp \"{O}ktem, Eric Paquin, Grace Tang, and Sylwia Tur. 2020.
\newblock {TICO}-19: the {T}ranslation initiative for {CO}vid-19.
\newblock {arXiv}:2007.01788.

\bibitem[{Axelrod et~al.(2011)Axelrod, He, and Gao}]{axelrod-etal-2011-domain}
Amittai Axelrod, Xiaodong He, and Jianfeng Gao. 2011.
\newblock Domain adaptation via pseudo in-domain data selection.
\newblock In \emph{Proceedings of EMNLP}.

\bibitem[{Britz et~al.(2017)Britz, Le, and Pryzant}]{britz-etal-2017-effective}
Denny Britz, Quoc Le, and Reid Pryzant. 2017.
\newblock Effective domain mixing for neural machine translation.
\newblock In \emph{Proceedings of the Second Conference on Machine
  Translation}.

\bibitem[{Chen et~al.(2020)Chen, Kornblith, Norouzi, and
  Hinton}]{pmlr-v119-chen20j}
Ting Chen, Simon Kornblith, Mohammad Norouzi, and Geoffrey Hinton. 2020.
\newblock A simple framework for contrastive learning of visual
  representations.
\newblock In \emph{Proceedings of ICLR}.

\bibitem[{Chu and Wang(2018)}]{chu-wang-2018-survey}
Chenhui Chu and Rui Wang. 2018.
\newblock A survey of domain adaptation for neural machine translation.
\newblock In \emph{Proceedings of COLING}.

\bibitem[{Cuong et~al.(2016)Cuong, Sima{'}an, and
  Titov}]{cuong-etal-2016-adapting}
Hoang Cuong, Khalil Sima{'}an, and Ivan Titov. 2016.
\newblock Adapting to all domains at once: Rewarding domain invariance in
  {SMT}.
\newblock \emph{TACL}, 4.

\bibitem[{Devlin et~al.(2019)Devlin, Chang, Lee, and
  Toutanova}]{devlin-etal-2019-bert}
Jacob Devlin, Ming-Wei Chang, Kenton Lee, and Kristina Toutanova. 2019.
\newblock {BERT}: Pre-training of deep bidirectional transformers for language
  understanding.
\newblock In \emph{Proceedings of NAACL-HTL}.

\bibitem[{Domhan and Hieber(2017)}]{domhan-hieber-2017-using}
Tobias Domhan and Felix Hieber. 2017.
\newblock Using target-side monolingual data for neural machine translation
  through multi-task learning.
\newblock In \emph{Proceedings of EMNLP}.

\bibitem[{Dou et~al.(2020)Dou, Anastasopoulos, and Neubig}]{Dou2020}
Zi-Yi Dou, Antonios Anastasopoulos, and Graham Neubig. 2020.
\newblock Dynamic data selection and weighting for iterative back-translation.
\newblock \emph{arXiv preprint arXiv:2004.03672}.

\bibitem[{Dou et~al.(2019)Dou, Hu, Anastasopoulos, and
  Neubig}]{dou-etal-2019-unsupervised}
Zi-Yi Dou, Junjie Hu, Antonios Anastasopoulos, and Graham Neubig. 2019.
\newblock Unsupervised domain adaptation for neural machine translation with
  domain-aware feature embeddings.
\newblock In \emph{Proceedings of EMNLP-IJCNLP}.

\bibitem[{Duh et~al.(2013)Duh, Neubig, Sudoh, and
  Tsukada}]{duh-etal-2013-adaptation}
Kevin Duh, Graham Neubig, Katsuhito Sudoh, and Hajime Tsukada. 2013.
\newblock Adaptation data selection using neural language models: Experiments
  in machine translation.
\newblock In \emph{Proceedings of ACL}.

\bibitem[{Farajian et~al.(2017)Farajian, Turchi, Negri, and
  Federico}]{farajian-etal-2017-multi}
M.~Amin Farajian, Marco Turchi, Matteo Negri, and Marcello Federico. 2017.
\newblock Multi-domain neural machine translation through unsupervised
  adaptation.
\newblock In \emph{Proceedings of the Second Conference on Machine
  Translation}.

\bibitem[{Ganin and Lempitsky(2015)}]{Ganin2015}
Yaroslav Ganin and Victor Lempitsky. 2015.
\newblock Unsupervised domain adaptation by backpropagation.
\newblock In \emph{Proceedings of the 32nd International Conference on
  International Conference on Machine Learning - Volume 37}, ICML’15, page
  1180–1189. JMLR.org.

\bibitem[{Gulcehre et~al.(2015)Gulcehre, Firat, Xu, Cho, Barrault, Lin,
  Bougares, Schwenk, and Bengio}]{gulcehre2015using}
Caglar Gulcehre, Orhan Firat, Kelvin Xu, Kyunghyun Cho, Loic Barrault, Huei-Chi
  Lin, Fethi Bougares, Holger Schwenk, and Yoshua Bengio. 2015.
\newblock On using monolingual corpora in neural machine translation.
\newblock \emph{arXiv preprint arXiv:1503.03535}.

\bibitem[{Han and Eisenstein(2019)}]{han-eisenstein-2019-unsupervised}
Xiaochuang Han and Jacob Eisenstein. 2019.
\newblock Unsupervised domain adaptation of contextualized embeddings for
  sequence labeling.
\newblock In \emph{Proceedings of the EMNLP-IJCNLP}.

\bibitem[{Hu et~al.(2019)Hu, Xia, Neubig, and Carbonell}]{hu-etal-2019-domain}
Junjie Hu, Mengzhou Xia, Graham Neubig, and Jaime Carbonell. 2019.
\newblock Domain adaptation of neural machine translation by lexicon induction.
\newblock In \emph{Proceedings of ACL}.

\bibitem[{Jin et~al.(2020)Jin, Jin, Tianyi~Zhou, and Szolovits}]{jin2020simple}
Di~Jin, Zhijing Jin, Joey Tianyi~Zhou, and Peter Szolovits. 2020.
\newblock A simple baseline to semi-supervised domain adaptation for machine
  translation.
\newblock \emph{arXiv e-prints}, pages arXiv--2001.

\bibitem[{Junczys-Dowmunt(2018)}]{junczys-dowmunt-2018-dual}
Marcin Junczys-Dowmunt. 2018.
\newblock Dual conditional cross-entropy filtering of noisy parallel corpora.
\newblock In \emph{Proceedings of the Third Conference on Machine Translation:
  Shared Task Papers}.

\bibitem[{Kingma and Ba(2015)}]{kingma:adam}
Diederick~P Kingma and Jimmy Ba. 2015.
\newblock Adam: A method for stochastic optimization.
\newblock In \emph{International Conference on Learning Representations
  (ICLR)}.

\bibitem[{Koehn et~al.(2007)Koehn, Hoang, Birch, Callison-Burch, Federico,
  Bertoldi, Cowan, Shen, Moran, Zens, Dyer, Bojar, Constantin, and
  Herbst}]{koehn-etal-2007-moses}
Philipp Koehn, Hieu Hoang, Alexandra Birch, Chris Callison-Burch, Marcello
  Federico, Nicola Bertoldi, Brooke Cowan, Wade Shen, Christine Moran, Richard
  Zens, Chris Dyer, Ond{\v{r}}ej Bojar, Alexandra Constantin, and Evan Herbst.
  2007.
\newblock {M}oses: Open source toolkit for statistical machine translation.
\newblock In \emph{Proceedings of the ACL: Demonstrations.}

\bibitem[{Koehn and Knowles(2017)}]{koehn-knowles-2017-six}
Philipp Koehn and Rebecca Knowles. 2017.
\newblock Six challenges for neural machine translation.
\newblock In \emph{Proceedings of WMT}.

\bibitem[{Kudo(2018)}]{kudo-2018-subword}
Taku Kudo. 2018.
\newblock Subword regularization: Improving neural network translation models
  with multiple subword candidates.
\newblock In \emph{Proceedings of the ACL}.

\bibitem[{Mahdieh et~al.(2020)Mahdieh, Chen, Cao, and Firat}]{Mahdieh2020}
Mahdis Mahdieh, Mia~Xu Chen, Yuan Cao, and Orhan Firat. 2020.
\newblock Rapid domain adaptation for machine translation with monolingual
  data.
\newblock \emph{arXiv preprint arXiv:2010.12652}.

\bibitem[{Mansour and Ney(2014)}]{mansour-ney-2014-unsupervised}
Saab Mansour and Hermann Ney. 2014.
\newblock Unsupervised adaptation for statistical machine translation.
\newblock In \emph{Proceedings of the Ninth Workshop on Statistical Machine
  Translation}.

\bibitem[{Moghimifar et~al.(2020)Moghimifar, Haffari, and
  Baktashmotlagh}]{moghimifar-etal-2020-domain}
Farhad Moghimifar, Gholamreza Haffari, and Mahsa Baktashmotlagh. 2020.
\newblock Domain adaptative causality encoder.
\newblock In \emph{Proceedings of the The 18th Annual Workshop of the
  Australasian Language Technology Association}.

\bibitem[{Moore and Lewis(2010)}]{moore-lewis-2010-intelligent}
Robert~C. Moore and William Lewis. 2010.
\newblock Intelligent selection of language model training data.
\newblock In \emph{Proceedings of ACL}.

\bibitem[{Obeid et~al.(2020)Obeid, Zalmout, Khalifa, Taji, Oudah, Alhafni,
  Inoue, Eryani, Erdmann, and Habash}]{obeid-etal-2020-camel}
Ossama Obeid, Nasser Zalmout, Salam Khalifa, Dima Taji, Mai Oudah, Bashar
  Alhafni, Go~Inoue, Fadhl Eryani, Alexander Erdmann, and Nizar Habash. 2020.
\newblock {CAM}e{L} tools: An open source python toolkit for {A}rabic natural
  language processing.
\newblock In \emph{Proceedings of LREC}.

\bibitem[{Ott et~al.(2019)Ott, Edunov, Baevski, Fan, Gross, Ng, Grangier, and
  Auli}]{ott-etal-2019-fairseq}
Myle Ott, Sergey Edunov, Alexei Baevski, Angela Fan, Sam Gross, Nathan Ng,
  David Grangier, and Michael Auli. 2019.
\newblock fairseq: A fast, extensible toolkit for sequence modeling.
\newblock In \emph{Proceedings of the NAACL: Demonstrations}.

\bibitem[{Parcheta et~al.(2018)Parcheta, Sanchis-Trilles, and
  Casacuberta}]{parcheta2018data}
Zuzanna Parcheta, Germ{\'a}n Sanchis-Trilles, and Francisco Casacuberta. 2018.
\newblock Data selection for nmt using infrequent n-gram recovery.

\bibitem[{Sanh et~al.(2019)Sanh, Debut, Chaumond, and
  Wolf}]{sanh2019distilbert}
Victor Sanh, Lysandre Debut, Julien Chaumond, and Thomas Wolf. 2019.
\newblock Distilbert, a distilled version of bert: smaller, faster, cheaper and
  lighter.
\newblock \emph{arXiv preprint arXiv:1910.01108}.

\bibitem[{Sennrich et~al.(2016)Sennrich, Haddow, and
  Birch}]{sennrich-etal-2016-improving}
Rico Sennrich, Barry Haddow, and Alexandra Birch. 2016.
\newblock Improving neural machine translation models with monolingual data.
\newblock In \emph{Proceedings of ACL}.

\bibitem[{Shah et~al.(2018)Shah, Lei, Moschitti, Romeo, and
  Nakov}]{shah-etal-2018-adversarial}
Darsh Shah, Tao Lei, Alessandro Moschitti, Salvatore Romeo, and Preslav Nakov.
  2018.
\newblock Adversarial domain adaptation for duplicate question detection.
\newblock In \emph{Proceedings of the EMNLP}.

\bibitem[{Shen et~al.(2018)Shen, Qu, Zhang, and Yu}]{Shen2018}
Jian Shen, Yanru Qu, Weinan Zhang, and Yong Yu. 2018.
\newblock Wasserstein distance guided representation learning for domain
  adaptation.
\newblock In \emph{Thirty-Second AAAI Conference on Artificial Intelligence}.

\bibitem[{Silva et~al.(2018)Silva, Liu, Poncelas, and
  Way}]{silva-etal-2018-extracting}
Catarina~Cruz Silva, Chao-Hong Liu, Alberto Poncelas, and Andy Way. 2018.
\newblock Extracting in-domain training corpora for neural machine translation
  using data selection methods.
\newblock In \emph{Proceedings of the Third Conference on Machine Translation:
  Research Papers}.

\bibitem[{Tiedemann(2012)}]{tiedemann-2012-parallel}
J{\"o}rg Tiedemann. 2012.
\newblock Parallel data, tools and interfaces in {OPUS}.
\newblock In \emph{Proceedings of {LREC}}.

\bibitem[{Vaswani et~al.(2017)Vaswani, Shazeer, Parmar, Uszkoreit, Jones,
  Gomez, Kaiser, and Polosukhin}]{transformer}
Ashish Vaswani, Noam Shazeer, Niki Parmar, Jakob Uszkoreit, Llion Jones,
  Aidan~N Gomez, \L~ukasz Kaiser, and Illia Polosukhin. 2017.
\newblock Attention is all you need.
\newblock In \emph{NIPS}.

\bibitem[{Vu et~al.(2020)Vu, Phung, and Haffari}]{vu-etal-2020-effective}
Thuy-Trang Vu, Dinh Phung, and Gholamreza Haffari. 2020.
\newblock Effective unsupervised domain adaptation with adversarially trained
  language models.
\newblock In \emph{Proceedings of the EMNLP}.

\bibitem[{Wang et~al.(2017)Wang, Finch, Utiyama, and
  Sumita}]{Wang2017SentenceEmbeddingNeural}
Rui Wang, Andrew Finch, Masao Utiyama, and Eiichiro Sumita. 2017.
\newblock Sentence embedding for neural machine translation domain adaptation.
\newblock In \emph{Proceedings of ACL}.

\bibitem[{Wang et~al.(2018)Wang, Qu, Chen, Shen, Zhang, Zhang, Gao, Gu, Chen,
  and Yu}]{wang-etal-2018-label-aware}
Zhenghui Wang, Yanru Qu, Liheng Chen, Jian Shen, Weinan Zhang, Shaodian Zhang,
  Yimei Gao, Gen Gu, Ken Chen, and Yong Yu. 2018.
\newblock Label-aware double transfer learning for cross-specialty medical
  named entity recognition.
\newblock In \emph{Proceedings of the NAACL}.

\bibitem[{Way et~al.(2018)Way, Poncelas, and Maillette~de
  Buy~Wenniger}]{way2018data}
Andy Way, Alberto Poncelas, and Gideon Maillette~de Buy~Wenniger. 2018.
\newblock Data selection with feature decay algorithms using an approximated
  target side.
\newblock IWSLT.

\end{thebibliography}
